\documentclass[tikz]{article}

\usepackage{PRIMEarxiv}

\usepackage[utf8]{inputenc} 
\usepackage[T1]{fontenc}    
\usepackage{hyperref}       
\usepackage{url}            
\usepackage{booktabs}       
\usepackage{amsfonts}       
\usepackage{nicefrac}       
\usepackage{microtype}      
\usepackage{lipsum}
\usepackage{fancyhdr}       
\usepackage{graphicx}       
\usepackage{csquotes}
\usepackage{mathtools} 
\usepackage{booktabs} 
\usepackage{algorithm}
\usepackage[noend]{algorithmic}
\usepackage{multirow}
\usepackage{caption}
\usepackage{subcaption}
\usepackage{amssymb}
\usepackage{graphicx}
\usepackage{xfrac}
\usepackage{xcolor}
\usepackage{ulem}
\usepackage{microtype}
\usepackage{hyperref}
\usepackage[acronym]{glossaries}
\usepackage{cleveref}

\title{
    PD-Loss: Proxy-Decidability for Efficient Metric Learning
}

\author{
Pedro Silva$^1$, Guilherme A. L. Silva$^1$, Pablo Coelho$^1$, Vander Freitas$^1$\\ 
\textbf{Gladston Moreira$^1$, David Menotii$^2$ and Eduardo Luz}$^1$ \\
 $^1$Computing Department, Universidade Federal de Ouro Preto, Ouro
Preto, 35402-136, Minas Gerais, Brazil.\\
$^1$Department of Informatics, Universidade Federal do Paraná, Curitba, 81531-990, Paraná, Brazil.\\
    \texttt{eduluz@ufop.edu.br}
}

\def\eg{\emph{e.g}.}

\begin{document}
\maketitle

\begin{abstract}
Deep Metric Learning (DML) aims to learn embedding functions that map semantically similar inputs to proximate points in a metric space while separating dissimilar ones. Existing methods, such as pairwise losses, are hindered by complex sampling requirements and slow convergence. In contrast, proxy-based losses, despite their improved scalability, often fail to optimize global distribution properties. The Decidability-based Loss (D-Loss) addresses this by targeting the decidability index ($d'$) to enhance distribution separability, but its reliance on large mini-batches imposes significant computational constraints. We introduce Proxy-Decidability Loss (PD-Loss), a novel objective that integrates learnable proxies with the statistical framework of $d'$ to optimize embedding spaces efficiently. By estimating genuine and impostor distributions through proxies, PD-Loss combines the computational efficiency of proxy-based methods with the principled separability of D-Loss, offering a scalable approach to distribution-aware DML. Experiments across various tasks, including fine-grained classification and face verification, demonstrate that PD-Loss achieves performance comparable to that of state-of-the-art methods while introducing a new perspective on embedding optimization, with potential for broader applications.
\end{abstract}

\keywords{Deep Metric Learning \and Proxy-based Loss \and Similarity Learning \and Computer Vision.}

\section{Introduction}
\label{sec:intro}
Learning effective distance metrics is paramount for many computer vision tasks. Metric learning focuses on learning powerful embedding functions, typically using deep neural networks, that project high-dimensional inputs such as images into a lower-dimensional space where semantic similarity directly translates into proximity according to a metric (\eg, Euclidean distance or cosine similarity). Well-learned embeddings have the potential to advance tasks such as instance retrieval~\cite{gordo2016deep}, face recognition~\cite{schroff_facenet_15}, person re-identification~\cite{hermans2017defense}, and few/zero-shot learning~\cite{snell2017prototypical}.

A dominant paradigm in DML involves losses operating on pairs, triplets, and quadriplets of samples within a mini-batch. Contrastive loss~\cite{hadsell2006dimensionality} pulls positive pairs together and pushes negative pairs apart.
Triplet loss~\cite{schroff_facenet_15, hermans2017defense} enforces a margin between the distance of an anchor to a positive sample and its distance to a negative sample. More advanced pairwise methods like N-Pair loss~\cite{sohn2016improved}, Multi-Similarity (MS) loss~\cite{wang_msloss_19}, and Circle loss~\cite{sun_circleloss_20} aim to improve sampling or gradient contribution by considering multiple pairs simultaneously or re-weighting pairs based on some criteria. While powerful, these methods often face significant challenges: their performance heavily depends on effective (and often computationally expensive) sample mining strategies (\eg, hard negative mining)~\cite{wu2017sampling}, careful tuning of hyperparameters like margins or scales, and they can suffer from slow convergence due to focusing only on a subset of relationships within a batch.

Proxy-based methods have been proposed to address the scalability and sampling issues of pairwise losses, since these methods can learn explicit representations (proxies) for each class in the embedding space. Losses like ProxyNCA~\cite{movshovitz2017no} or ProxyAnchor~\cite{kim2020proxy} optimize the distance between samples and proxy vectors, eliminating the need for pair/triplet mining within a batch, which leads to faster convergence and better scalability to datasets with a large number of classes. 
However, the optimization target is indirect: optimizing sample-proxy distances does not explicitly guarantee optimal separation between all positive and negative pairs from different classes, and performance can be sensitive to the initialization and quality of the proxies.

An alternative perspective was offered by the Decidability Loss (D-Loss)~\cite{SilvaIJCNN22}, inspired by the Decidability Index ($d'$) from biometrics~\cite{daugman2000biometric}. 
D-Loss directly optimizes the separability of the distributions of distances (or similarities) between genuine pairs (in the same class) and impostor pairs (different classes) within a mini-batch. Its goal is to maximize the distance between the means of these two distributions while minimizing their variances simultaneously. This provides a global, statistical view of embedding quality, is parameter-free (no margins to tune), and directly targets distribution separability. However, its formulation relies heavily on calculating these statistics from all pairs within a mini-batch, which requires impractically large batch sizes for stable estimates and makes it sensitive to outliers within a batch~\cite{SilvaIJCNN22}.
\begin{figure}[!ht]
    \centering
    \vspace{-3mm}
    \includegraphics[width=\linewidth]{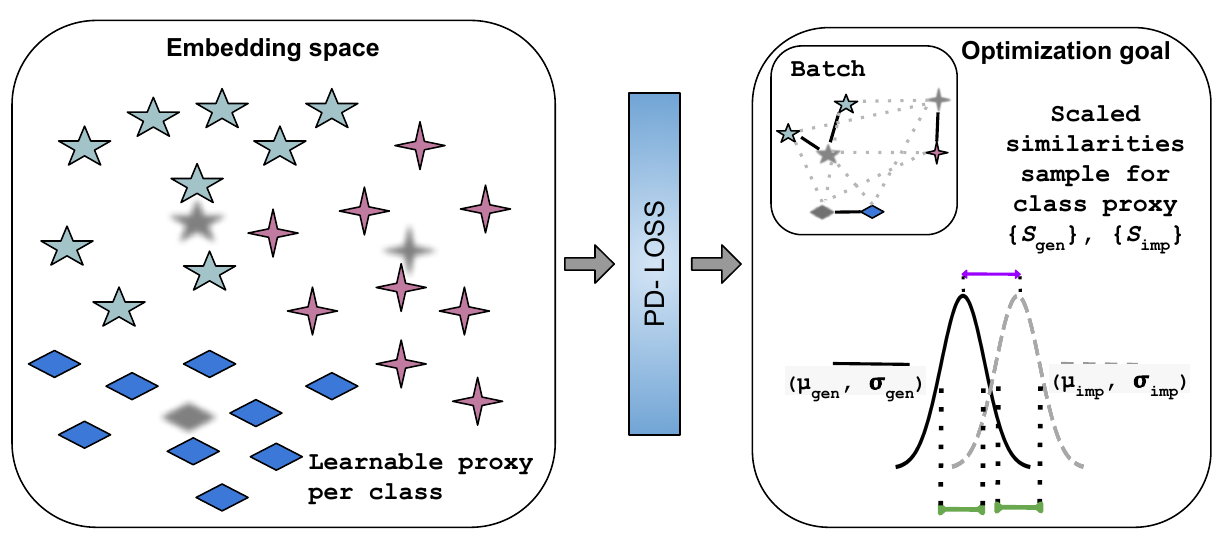}
    \caption{Conceptual illustration of PD-Loss.
    Left: Samples from different classes ($\textcolor{cyan}{\star}, \textcolor{magenta}{+}, \textcolor{blue}{\diamond}$) and their corresponding learnable proxies ($\textcolor{black}{\star}, \textcolor{black}{+}, \textcolor{black}{\diamond}$) co-exist in the embedding space.
    Center: Scaled similarities between samples and proxies are computed. These similarities form the input for the PD-Loss function.
    Right: The PD-Loss objective aims to maximize the Decidability Index ($d'$) derived from the estimated genuine (sample-to-correct-proxy) and impostor (sample-to-incorrect-proxy) similarity distributions. This involves maximizing the separation between their means ($\mu_{gen}, \mu_{imp}$) and minimizing their variances ($\sigma^2_{gen}, \sigma^2_{imp}$). Both the backbone network and the proxies are updated via gradient descent. }
    \vspace{-3mm}   
\label{fig:pd_loss_concept}
\end{figure}

This paper introduces Proxy-Decidability Loss (PD-Loss), a novel approach that retains D-Loss's core statistical optimization principle while overcoming its batch size dependency. We propose using learnable class proxies, similar to proxy-based methods, solely for estimating the genuine and impostor similarity distributions required to compute the Decidability Index. Specifically, the genuine distribution is estimated from sample-to-correct-proxy similarities, and the impostor distribution from sample-to-incorrect-proxy similarities within a mini-batch. We further enhance stability and performance by optimizing the logarithm of the inverse Decidability Index and incorporating temperature scaling. We hypothesize that by combining the efficient distribution estimation enabled by proxies with the statistically grounded optimization objective of the Decidability Index, PD-Loss can achieve strong DML performance while being more robust to batch size variations and potentially requiring less hyperparameter tuning compared to margin-based pairwise losses. Figure \ref{fig:pd_loss_concept} provides a conceptual illustration of PD-Loss.

We validate PD-Loss through extensive experiments on three challenging benchmark datasets: CUB-200-2011~\cite{wah2011caltech} and Stanford Cars~\cite{krause20133d} for fine-grained visual categorization, and Labeled Faces in the Wild (LFW)~\cite{LFWTechUpdate} for face verification. We compare PD-Loss against state-of-the-art baselines including ProxyNCA~\cite{movshovitz2017no}, ProxyAnchor~\cite{kim2020proxy}, Multi-Similarity Loss~\cite{wang_msloss_19}, and Circle Loss~\cite{sun_circleloss_20}. Our results demonstrate that PD-Loss achieves highly competitive results. Ablations reveal that random proxy initialization is also effective in our setup, highlighting PD-Loss's robustness to variations in batch size, a key advantage over pair-based methods. These findings confirm that optimizing distribution separability via proxy-based estimation is a viable strategy for deep metric learning. The source code for this paper is available in \url{https://github.com/anonimous_due_to_peer_review}.

\section{Background}
\label{sec:background}

This section outlines the foundational concepts relevant to this work. 

\subsection{Problem Setting and Notation}
Deep Metric Learning (DML) addresses the task of learning an embedding function $f: \mathcal{X} \rightarrow \mathcal{Z} \subseteq \mathbb{R}^D$, which maps samples $x$ from an input space $\mathcal{X}$ (e.g., images) to a $D$-dimensional embedding space $\mathcal{Z}$. The goal is to structure $\mathcal{Z}$ such that the distance $d(f(x_i), f(x_j))$ (or similarity $s(f(x_i), f(x_j))$) between embeddings reflects the semantic similarity of the original inputs $x_i$ and $x_j$. Given a dataset $\mathcal{D} = \{(x_i, y_i)\}_{i=1}^N$, where $y_i \in \{1, ..., C\}$ is the class label for sample $x_i$ out of $C$ classes, the objective is to learn $f$ such that embeddings $z_i = f(x_i)$ and $z_j = f(x_j)$ are close if $y_i = y_j$ (genuine/positive pair) and far apart if $y_i \neq y_j$ (impostor/negative pair). We primarily consider the cosine similarity $s(z_i, z_j) = \frac{z_i \cdot z_j}{\|z_i\| \|z_j\|}$ after L2-normalizing embeddings, making the cosine distance $d_{cos}(z_i, z_j) = 1 - s(z_i, z_j)$.

\subsection{Pairwise and Triplet Losses}

A common DML strategy involves losses defined over pairs or triplets of samples within a mini-batch $B$. The Triplet Loss~\cite{schroff_facenet_15} is a prominent example. For a triplet $(z_a, z_p, z_n)$ consisting of an anchor $z_a$, a positive sample $z_p$ ($y_a=y_p$), and a negative sample $z_n$ ($y_a \neq y_n$), the loss enforces a margin $\alpha$:
\begin{equation}
    \mathcal{L}_{Triplet} = \sum_{(a,p,n) \in B} \max(0, d(z_a, z_p)^2 - d(z_a, z_n)^2 + \alpha),
    \label{eq:triplet}
\end{equation}
where $d(\cdot, \cdot)$ is often the Euclidean distance. Multi-Similarity Loss (MS-Loss)~\cite{wang_msloss_19} and Circle Loss~\cite{sun_circleloss_20} refine this by considering all positive and negative pairs relative to an anchor and re-weighting them based on their similarity, aiming to pull hard positives closer and push hard negatives further, more effectively than the triplet loss. These methods rely heavily on informative pairs present in the mini-batch or require sophisticated mining strategies~\cite{wu2017sampling}.

\subsection{Proxy-Based Losses}
Proxy-based methods learn a set of $C$ proxy vectors $\mathcal{P} = \{p_c\}_{c=1}^C$, where each $p_c \in \mathbb{R}^D$ represents class $c$. Instead of comparing samples to each other, these methods compare samples to proxies. Proxy-NCA~\cite{movshovitz2017no} uses a Softmax-like loss, encouraging each sample $z_i$ (with label $y_i$) to be closer to its corresponding proxy $p_{y_i}$ than to any other proxy $p_k$ ($k \neq y_i$):
\begin{equation}
    \mathcal{L}_{ProxyNCA} = \sum_{i} -\log \left( \frac{\exp(-d(z_i, p_{y_i}))}{\sum_{k \neq y_i} \exp(-d(z_i, p_k))} \right).
    \label{eq:proxynca} 
\end{equation}
ProxyAnchor~\cite{kim2020proxy} refines this by treating each proxy as an anchor and applying a margin-based loss using all samples in the batch as positives or negatives relative to that proxy, offering improved gradient signals. While scalable, the connection between sample-proxy distances and the desired structure of sample-sample distances is indirect.

\subsection{Decidability Index and D-Loss}
The Decidability Index ($d'$)~\cite{daugman2000biometric} quantifies the separation between two distributions, typically the genuine (intra-class) and impostor (inter-class) score distributions in verification systems. Given the means ($\mu_g, \mu_i$) and standard deviations ($\sigma_g, \sigma_i$) of the genuine and impostor distributions respectively, $d'$ is defined as:
\begin{equation}
    d' = \frac{|\mu_i - \mu_g|}{\sqrt{(\sigma_g^2 + \sigma_i^2) / 2}}.
    \label{eq:dprime}
\end{equation}
A higher $d'$ indicates better separability. The D-Loss function, presented in \cite{SilvaIJCNN22}, proposes minimizing its inverse, using statistics derived from all genuine and impostor pairs within a mini-batch:
\begin{equation}
    \mathcal{L}_{D\mbox{-}Loss} = \frac{1}{d'_{batch}} = \frac{\sqrt{(\sigma_{g,batch}^2 + \sigma_{i,batch}^2) / 2}}{|\mu_{i,batch} - \mu_{g,batch}| + \epsilon}.
    \label{eq:dloss_original}
\end{equation}
This directly optimizes the statistical properties of the embedding space but requires large batches for reliable estimation of the batch statistics ($\mu_{batch}, \sigma_{batch}$). 

\section{Proxy-Decidability Loss (PD-Loss)}
\label{sec:method}

The original D-Loss~\eqref{eq:dloss_original} offers an appealing objective by directly optimizing the statistical separability ($d'$) of genuine and impostor score distributions. However, its reliance on mini-batch statistics makes it impractical for standard batch sizes and susceptible to noise. Proxy-based losses, on the other hand, offer scalability by avoiding pair sampling. We propose PD-Loss to combine the best of both worlds: leveraging proxies for efficient distribution estimation while optimizing the robust $d'$ objective.

Let $\mathcal{P} = \{p_c \in \mathbb{R}^D\}_{c=1}^C$ be the set of learnable proxies, one for each of the $C$ classes. Those proxies could be precomputed, such as the mean embeddings of instances of a given class, or could be initialized randomly.
\begin{equation}
    s_{i,c} = (\tilde{z}_i^T \tilde{p}_c) / \tau.
    \label{eq:scaled_similarity}
\end{equation}
Here, $s_{i,c}$ denotes the scaled similarity between the embedding of sample $i$ and the proxy for class $c$. Instead of computing distances between all pairs $(z_i, z_j)$ in the batch, we use the sample-proxy similarities $s_{i,c}$ to estimate the genuine and impostor distributions relative to the proxies. For each sample $(x_i, y_i)$ in the mini-batch $B$, we consider its scaled similarity with its correct proxy ($p_{y_i}$) and with all incorrect proxies ($p_c$ where $c \neq y_i$). $\tau$ is the temperature.

\textbf{Genuine Similarities ($S_{gen}$):} This set collects the scaled similarities between each sample $i$ in the batch and its corresponding class proxy $p_{y_i}$: $S_{gen} = \{ s_{i, y_i} \mid (x_i, y_i) \in B \}$. The size of $S_{gen}$ is $|B|$, the number of samples in the mini-batch. Each element $s_{i, y_i}$ represents the scaled similarity of sample $i$ with its genuine proxy $p_{y_i}$.

\textbf{Impostor Similarities ($S_{imp}$):} This set collects the scaled similarities between each sample $i$ in the batch and all proxies $p_c$ belonging to classes different from the sample's class $y_i$: $S_{imp} = \{ s_{i, c} \mid (x_i, y_i) \in B, c \neq y_i \}$. The size of $S_{imp}$ is $|B| \times (C-1)$, where $C$ is the total number of classes. Each element $s_{i, c}$ (where $c \neq y_i$) represents the scaled similarity of sample $i$ with an impostor proxy $p_c$.

These sets $S_{gen}$ and $S_{imp}$ provide the empirical distributions of scaled genuine and impostor similarities observed within the current mini-batch, based on interactions with the learned class proxies. We compute the statistics (mean $\mu$, variance $\sigma^2$) from these estimated distributions: $\mu_{gen} = \text{mean}(S_{gen})$, $\sigma^2_{gen} = \text{var}(S_{gen})$, $\mu_{imp} = \text{mean}(S_{imp})$, and $\sigma^2_{imp} = \text{var}(S_{imp})$.

Our goal is to maximize the separability, which for similarities means maximizing $\mu_{gen}$ relative to $\mu_{imp}$ while minimizing variances. We adapt the inverse $d'$ formulation, using similarity statistics and aiming for maximization (equivalent to minimizing negative $d'$). For numerical stability and potentially better gradient properties, we optimize the log-negative-$d'$ formulation (similar to Equation~\eqref{eq:dloss_original} but adapted for similarity and using log):
\begin{equation}
    \mathcal{L}_{PD} = -\log(\mu_{gen} - \mu_{imp} + \epsilon_1) + \frac{1}{2}\log(\sigma^2_{gen} + \sigma^2_{imp} + \epsilon_2)
    \label{eq:pd_loss}
\end{equation}
where $\epsilon_1, \epsilon_2$ are small positive constants (e.g., $10^{-6}$) for numerical stability, ensuring the arguments to the logarithms remain positive. Minimizing $\mathcal{L}_{PD}$ pushes $\mu_{gen}$ higher than $\mu_{imp}$ (first term) and simultaneously encourages smaller variances $\sigma^2_{gen}$ and $\sigma^2_{imp}$ (second term).

The initial state of the proxies $p_c$ can influence convergence. While random initialization (e.g., Kaiming uniform) is possible, the models could benefit from a more informed starting point. We also investigate a pre-computation strategy: before the first training epoch, we perform a forward pass over the entire training set using the initial (pretrained) backbone, calculate the mean L2-normalized embedding for each class, and use these mean embeddings as the initial values for the proxies $p_c$. 

The backbone network parameters $\theta_f$ and the proxy parameters $\mathcal{P}$ are optimized jointly. We compute the gradients of $\mathcal{L}_{PD}$~\eqref{eq:pd_loss} concerning both sets of parameters using standard backpropagation and update them using an optimizer like AdamW~\cite{loshchilov2017decoupled}. We optionally employ a higher learning rate for the proxies compared to the backbone, as proxies might need to adapt more quickly initially. Gradient clipping is also optionally applied for stability.

PD-Loss aims to combine the strengths of D-Loss and proxy-based methods. Using proxies for estimation avoids the $O(N^2)$ complexity of intra-batch pair calculation required by the original D-Loss and typical pairwise losses, making it efficient and less dependent on large batch sizes. Unlike standard proxy losses that optimize sample-proxy distances directly, PD-Loss uses proxies to estimate global distribution statistics (mean, variance) and optimizes the Decidability Index, a measure directly related to class separability. Furthermore, like D-Loss, it avoids explicit margin hyperparameters common in triplet or Circle loss. From this point onward, we use BS to denote Batch Size and EMB to denote Embedding Size.

\section{Experiments}
\label{sec:experiments}

We evaluate the PD-Loss and compare it against state-of-the-art DML baselines on three diverse benchmark datasets.
All experiments are implemented in PyTorch~\cite{paszke2019pytorch}. We use standard ImageNet-pretrained ResNet50~\cite{he2016deep} as the backbone. The final classification layer is replaced by a linear layer projecting to the embedding dimension $D$, followed by L2 normalization. We use the AdamW optimizer~\cite{loshchilov2017decoupled} with a base learning rate of $1 \times 10^{-5}$ and weight decay of $1 \times 10^{-4}$. A cosine annealing learning rate scheduler is used over 500 epochs for the main comparison and key ablations. Batch size is 32 for the main comparison unless otherwise specified. For PD-Loss and ProxyNCA, proxies are learnable parameters. Gradient clipping with max norm 1.0 is applied. Baseline hyperparameters are tuned based on their respective papers or validation set performance. Experiments are run on NVIDIA RTX 3090 GPUs. 

\subsection{Datasets}
    The \textbf{CUB-200-2011 (CUB)}~\cite{wah2011caltech} is a fine-grained bird classification dataset with 200 classes. We use the standard split of 5,994 training images and 5,794 test images. We further split 10\% of the training data for validation. The \textbf{Stanford Cars (CARS196)}~\cite{krause20133d} is a Fine-grained car classification dataset with 196 classes. Contains 8,144 training images and 8,041 test images. We use the training set and optionally reserve a small fraction (e.g., 20\%) for validation, using the labeled test set for final evaluation. Finally, the \textbf{Labeled Faces in the Wild (LFW)}~\cite{LFWTechUpdate} is a standard face verification benchmark. We use the LFWPeople subset via scikit-learn's lib, filtering for identities with at least 40 images (min faces per person = 40), resulting in 19 classes suitable for metric learning evaluation. We use the standard train/test splits defined by the loader or create internal validation splits. The choice of datasets covers fine-grained recognition (CUB, CARS) and a standard biometric verification task (LFW).

\vspace{-2mm}

\subsection{D-Loss vs. PD-Loss}
\begin{figure}[!htb]
    \centering
    \includegraphics[width=0.5\textwidth]{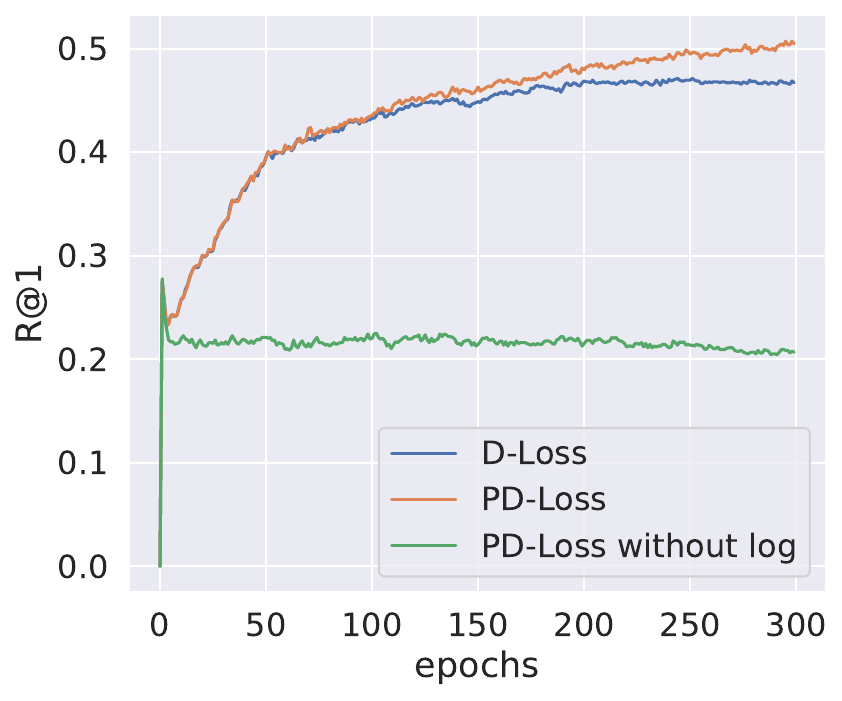}
    \caption{Comparison against the D-Loss, PD-Loss without the logarithmic term, and with it. The R@1 is calculated over the epochs in the validation split of CUB200 dataset. We used a batch size of 500, the ResNet-18 architecture, and the embedding size of 256.}
    \vspace{-1mm}
    \label{fig:dloss-pdlossbel}
\end{figure}

In Figure~\ref{fig:dloss-pdlossbel}, we present the Recall@1 (R@1) performance over 300 training epochs on the validation split of the CUB200 dataset. 
In order to conduct experiments with a greater batch size, we kept the ResNet architecture. Still, instead of using the ResNet50, we adopted the ResNet-18, embedding layer comprising 256 units and a batch size of 500. These hyperparameters closely align with those utilized in the original D-Loss formulation proposed by Silva et al.~\cite{SilvaIJCNN22}, ensuring a fair and consistent basis for comparison.
As illustrated in Figure~\ref{fig:dloss-pdlossbel}, the proposed PD-Loss consistently outperforms the D-Loss in terms of R@1. Notably, the inclusion of the logarithmic term in PD-Loss plays a critical role in enabling effective model training, indicating its importance in shaping the loss behaviour and guiding optimization. A significant advantage of PD-Loss over D-Loss lies in its flexibility with respect to batch size. While D-Loss relies on the presence of genuine pairs (and impostors) within each training batch to construct a representative distribution, this assumption becomes increasingly difficult to satisfy with smaller batch sizes and datasets with a greater number of classes. In contrast, PD-Loss alleviates this limitation, allowing the model to be trained effectively even in low-batch regimes.

In summary, PD-Loss constitutes a robust extension of D-Loss, offering improved retrieval performance (as evidenced by higher R@1 scores) while simultaneously enhancing practical applicability through its ability to operate effectively with smaller batch sizes.

\subsection{Baselines and Evaluation Metrics}
We compare PD-Loss (our proposed method) against the following representative DML losses: (i) \textbf{ProxyNCA}~\cite{movshovitz2017no} - A standard proxy-based loss using NCA objective, (ii) \textbf{ProxyAnchor}~\cite{kim2020proxy} - An improved proxy-based loss with better gradient properties, (iii) \textbf{Multi-Similarity (MS) Loss}~\cite{wang_msloss_19} - A strong pair-based loss considering multiple similarities that usually requires a miner and (iv) \textbf{Circle Loss}~\cite{sun_circleloss_20} - A unified pair-based loss formulation aiming for more flexible optimization boundaries, that also requires a miner. For MS Loss and Circle Loss, we employ suitable miners like MultiSimilarityMiner or PairMarginMiner from PyTorch metric learning~\cite{musgrave2020pytorch}. Hyperparameters for all baselines are tuned based on runs on the training dataset.

\begin{figure}[!ht]
    \centering
        \includegraphics[width=0.5\linewidth]{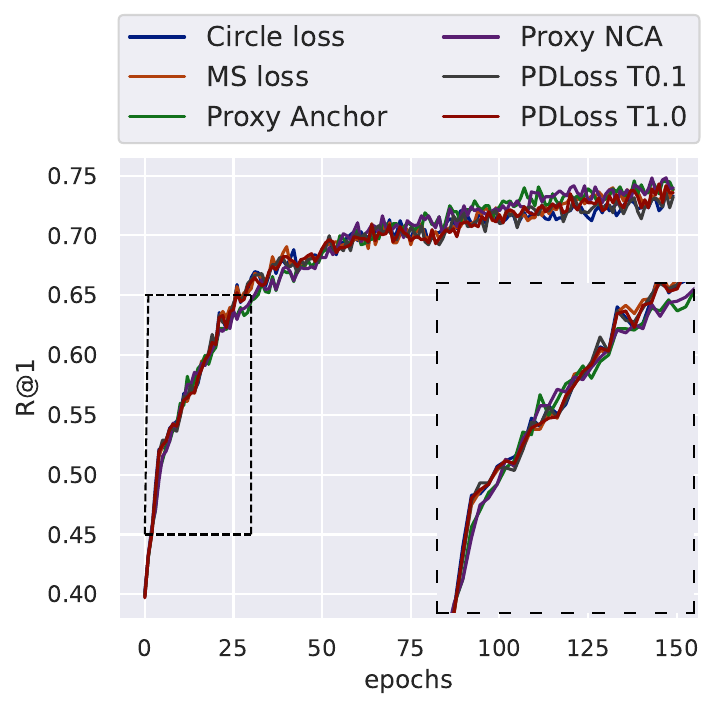}
    \caption{Presents a comparison of different losses in terms of R@1 over training epochs.}
    \label{fig:Loss_epochs}
\end{figure}

This work primarily uses Recall@K (R@K) for K={1, 2, 4, 8} as the evaluation metric, standard in retrieval and recognition tasks. R@K measures the proportion of queries for which at least one correct match is found within the top K retrieved neighbours. We calculate neighbours based on the cosine distance between L2-normalized embeddings. Figure~\ref{fig:Loss_epochs} shows the training result of all techniques in 150 epochs. The original D-Loss~\cite{SilvaIJCNN22} is not included as a baseline in our experiments because its reliance on large mini-batches for stable statistic estimation makes its performance inconsistent with the standard batch sizes used here.

\subsection{Ablation Studies}
\label{sec:ablation_studies}

We conduct comprehensive ablation studies on the CUB-200 validation set (EMB 512, BS 32, 500 epochs) to understand the impact of critical design choices and hyperparameters in PD-Loss.
\begin{figure}[!ht]
    \centering
        \includegraphics[width=3.2in,height=2.2in]{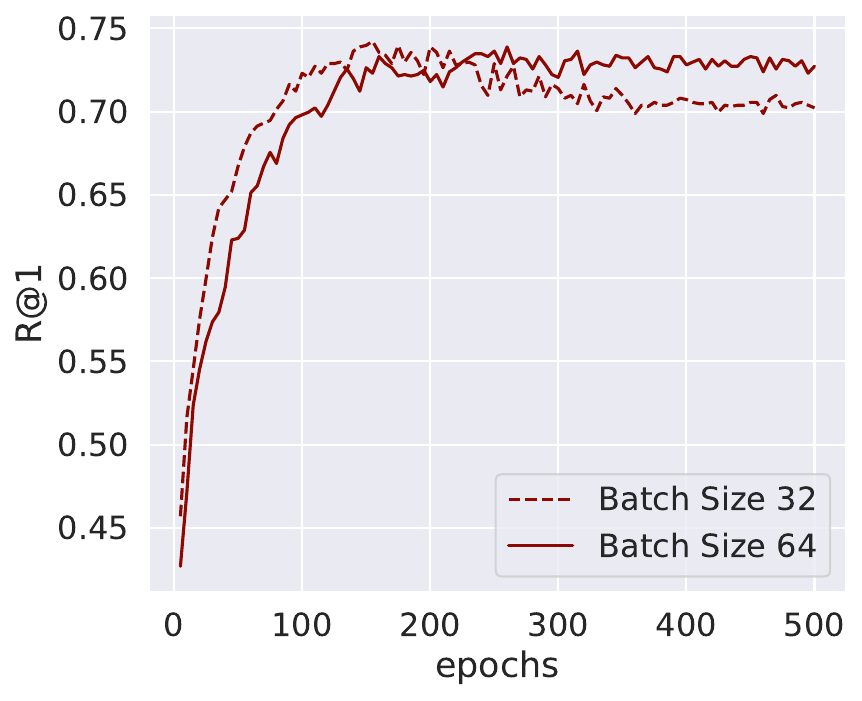 }
    \caption{Presents the analysis of the batch size impact in PD-Loss (Random Init $\tau=1.0$) on CUB-200 val set (EMB 512, 500 epochs), with R@1 curves for BS$=\{32, 64\}$.}
    \label{fig:ablation_batchsize}
\end{figure} 

\subsubsection{Batch Size (BS) Sensitivity}
\label{sec:ablation_batchsize}

We analyze the impact of batch size on PD-Loss convergence and final performance, using the configuration {EMB 512, $\tau=1.0$, Random Init}. Figure~\ref{fig:ablation_batchsize} plots validation Recall@1 across training epochs for batch sizes 32 and 64. While BS 32 converges slightly faster initially, BS 64 exhibits greater stability in the later stages of training and achieves a slightly higher final R@1. This suggests that while PD-Loss is not as heavily dependent on huge batches as the original D-Loss, using moderately larger batches can still provide more stable gradient estimates, benefiting convergence behavior.

\subsubsection{Proxy Initialization and Temperature ($\tau$)}
\label{sec:ablation_init_temp}
We evaluate the combined effect of the proxy initialization strategy (Random vs. Precomputed) and the temperature parameter $\tau$. 
\begin{table*}[!ht]
    \centering
    \caption{Ablation study on proxy initialization strategy and temperature $\tau$ for PD-Loss on CUB-200 validation set (EMB 512, BS 32, 500 epochs).}
    \label{tab:ablation_init_temp}
    \resizebox{\linewidth}{!}{
        \begin{tabular}{l@{\extracolsep{2mm}}c@{\extracolsep{0mm}\hskip3mm}c@{\hskip3mm}c@{\hskip3mm}c@{\extracolsep{2mm}}c@{\extracolsep{0mm}\hskip3mm}c@{\hskip3mm}c@{\hskip3mm}c@{\extracolsep{2mm}}c@{\extracolsep{0mm}\hskip3mm}c@{\hskip3mm}c@{\hskip3mm}c}
            \toprule[2pt]        
            \multirow{2}{*}{Initialization Type}
                & \multicolumn{4}{c}{PD-Loss $\tau=0.1$}
                & \multicolumn{4}{c}{PD-Loss $\tau=0.5$}
                & \multicolumn{4}{c}{PD-Loss $\tau=1.0$}\\\cmidrule{2-5} \cmidrule{6-9} \cmidrule{10-13}
                
                & \parbox{1cm}{\centering R@1}
                & \parbox{1cm}{\centering R@2}
                & \parbox{1cm}{\centering R@4}
                & \parbox{1cm}{\centering R@8}
                & \parbox{1cm}{\centering R@1}
                & \parbox{1cm}{\centering R@2}
                & \parbox{1cm}{\centering R@4}
                & \parbox{1cm}{\centering R@8}
                & \parbox{1cm}{\centering R@1}
                & \parbox{1cm}{\centering R@2}
                & \parbox{1cm}{\centering R@4}
                & \parbox{1cm}{\centering R@8}\\ \midrule
            Random Init 
                & 0.7629 & 0.8229 & 0.8664 & 0.8975 
                & 0.7784 & 0.8386 & 0.8785 & 0.9075
                & 0.7794 & 0.8352 & 0.8757 & 0.9104 \\
            Precomputed Init
                & 0.7684 & 0.8290 & 0.8726 & 0.9016 
                & 0.7742 & 0.8371 & 0.8802 & 0.9063
                & 0.7682 & 0.8340 & 0.8761 & 0.9052 \\
            \bottomrule[2pt]
        \end{tabular}
    }
\end{table*}
As shown in Table~\ref{tab:ablation_init_temp}, random proxy initialization with $\tau=1.0$ achieved the best R@1 (0.7794) among these variants. 
Comparing initialization strategies, random initialization performed better than precomputed means for both $\tau=0.1$ and $\tau=1.0$. This suggests that allowing the proxies full flexibility to learn their positions from a random start is more beneficial than constraining them to cluster centers derived from the initial, potentially sub-optimal, pretrained feature space.
Comparing temperature values within random initialization, $\tau=1.0$ (R@1 0.7794) outperformed $\tau=0.1$ (R@1 0.7629). This indicates that a higher temperature, which increases the sharpness of similarity distributions, could be better for achieving better performance with PD-Loss in this setting. Based on these findings, we use random proxy initialization and Temperature $\tau=1.0$ as the selected configuration for PD-Loss in the main comparison. It is observable in the remaining metrics (R@2, R@4 and R@8).

\subsubsection{Embedding Size (EMB) Impact}
\label{sec:ablation_emb_size}

Table~\ref{tab:ablation_emb_size} summarizes the impact of different embedding sizes for PD-Loss (Random Init $\tau=1.0$, BS 32, 500 epochs). EMB 512 achieves the highest R@1, although 1024 yields slightly better R@4 and R@8. This suggests that increasing embedding size beyond 512 in this specific setting does not linearly translate to R@1 gains but might benefit retrieval at higher ranks. We use 512 as the default embedding size for the main comparison, balancing performance and dimensionality.
\begin{table}[!ht]
    \centering
    \caption{Ablation study on embedding size for PD-Loss (Random Init $\tau=1.0$, BS 32, 500 epochs) on CUB-200 validation set.}
    \label{tab:ablation_emb_size}
        \begin{tabular}{l@{\hskip3mm}c@{\hskip3mm}c@{\hskip3mm}c@{\hskip3mm}c}
            \toprule[2pt]        
            Embedding Size (EMB)
                & \parbox{1cm}{\centering R@1}
                & \parbox{1cm}{\centering R@2}
                & \parbox{1cm}{\centering R@4}
                & \parbox{1cm}{\centering R@8} \\ \midrule
            256 & 0.7641 & 0.8381 & 0.8964 & 0.9355 \\ 
            512 & 0.7794 & 0.8352 & 0.8757 & 0.9104 \\ 
            1024 & 0.7635 & 0.8428 & 0.9021 & 0.9367 \\ 
            \bottomrule[2pt]
        \end{tabular}    
\end{table}

\subsubsection{Main Comparison} 
\label{sec:main_comparison}
%
Based on the ablation studies, we select PD-Loss with random proxy initialization and Temperature $\tau=1.0$ as our best variant. We compare this configuration against the baselines (ProxyNCA, ProxyAnchor, MS Loss, Circle Loss) on CUB-200, CARS196, and LFW using the configuration {EMB 512, BS 32, 500 epochs}.
\begin{figure}[!ht]
    \centering
    \begin{tabular}{@{}c@{}c@{}c@{ }c@{}}
        \includegraphics[width=1.68in,height=1.65in]{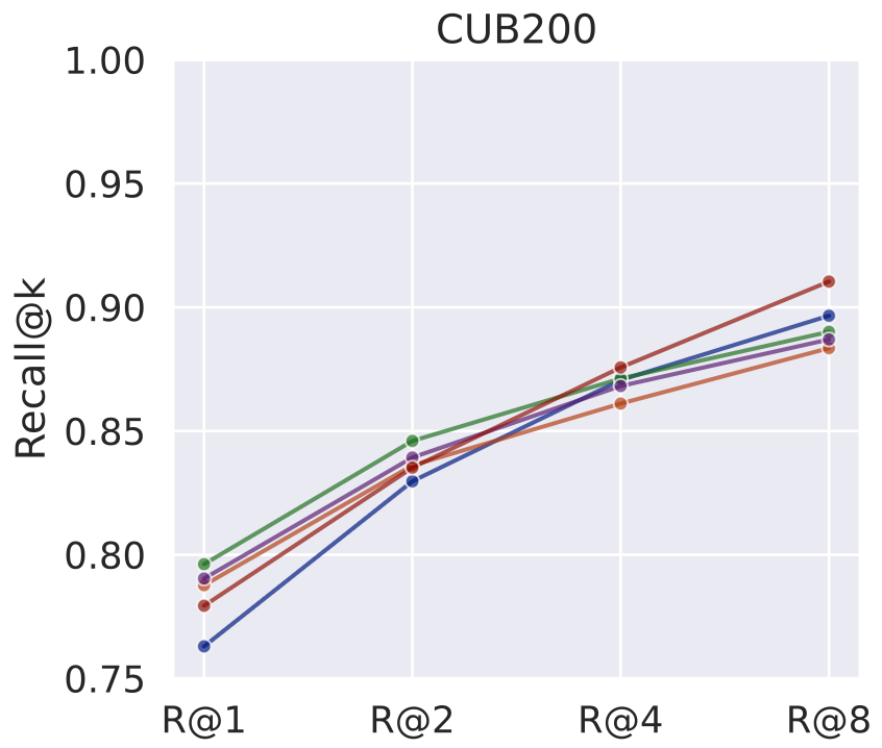} & 
        \includegraphics[width=1.68in,height=1.65in]{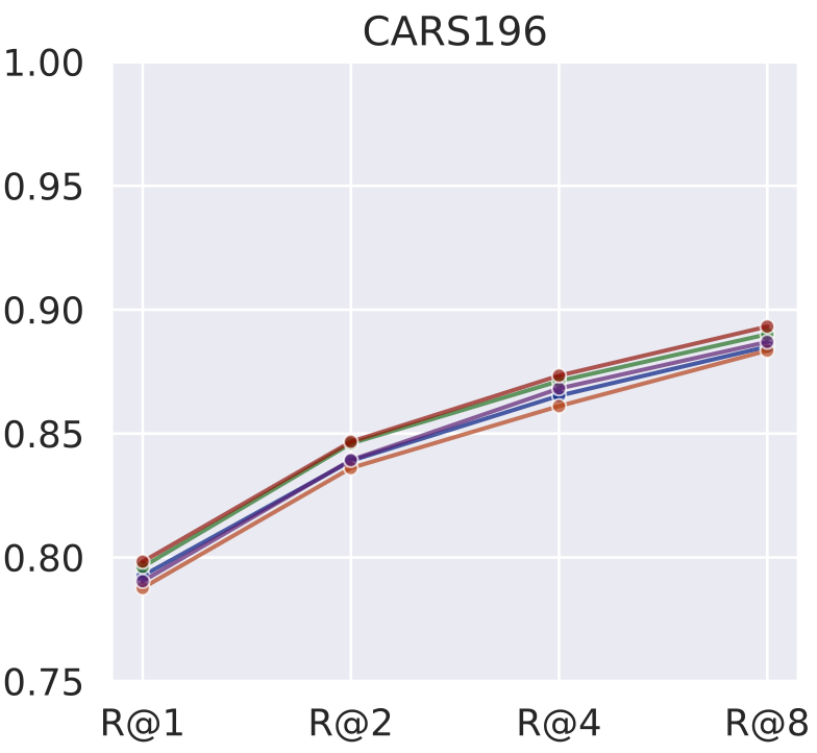 } &
        \includegraphics[width=1.68in,height=1.65in]{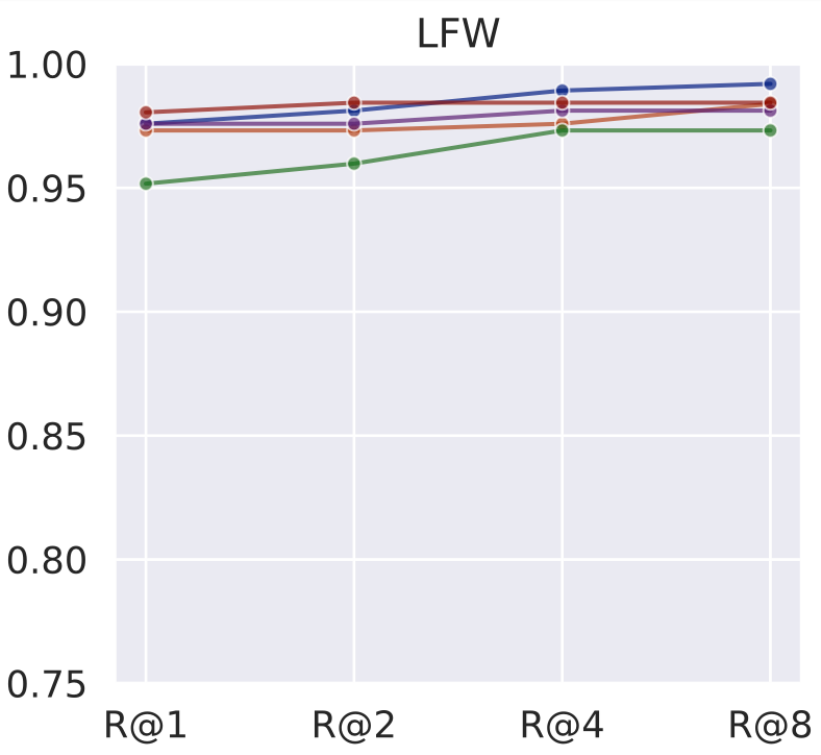 } &
        \includegraphics[width=1.68in,height=1.65in]{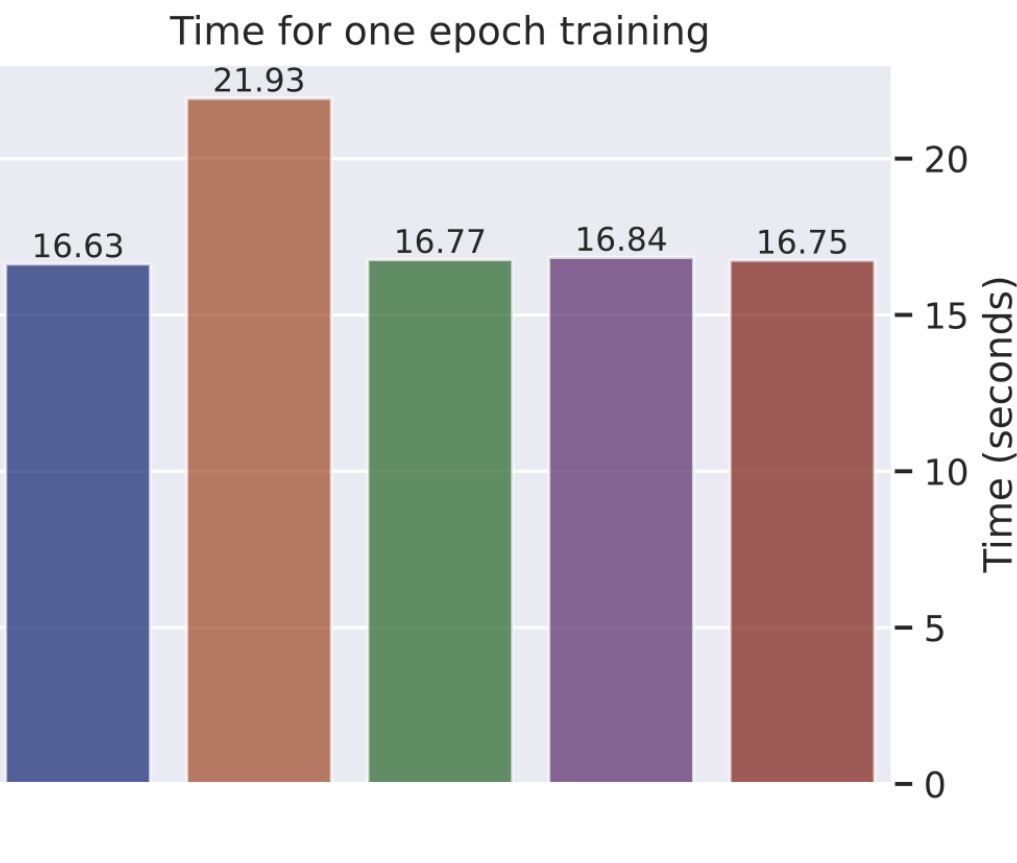 } \\
        (a)& (b)&(c)& (d)\\
    \end{tabular}
    \includegraphics[width=.75\linewidth]{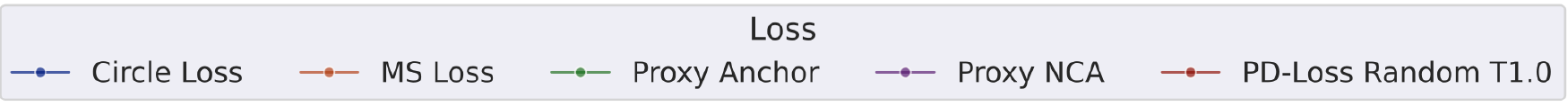 }
\caption{PD-Loss Random $\tau=1.0$ is compared against baselines: Comparative Recall@K performance on (a) CUB-200, (b) CARS196, and (c) LFW test sets. (d) Average training time per epoch.}
     \label{fig:main_comparison}
\end{figure}

Figure~\ref{fig:main_comparison} presents a comparative analysis of PD-Loss Random T1.0 against Circle Loss, MS Loss, Proxy Anchor, and Proxy NCA across three benchmark datasets: CUB-200, CARS196, and LFW.
In Figure~\ref{fig:main_comparison} (a), which shows results on the CUB-200 dataset, PD-Loss Random T1.0 achieved the highest R@4 and R@8. Figure~\ref{fig:main_comparison} (b), corresponding to CARS196, reveals a similar trend: PD-Loss leads across all metrics. Figure~\ref{fig:main_comparison} (c) reports results on LFW, where it can be seen that all methods perform well (R@1 $>$ 0.95); nonetheless, PD-Loss remains highly competitive, ranking among the top methods, achieving the highest R@1 and R@2, and reaching the second highest R@4 and R@8. Finally, Figure~\ref{fig:main_comparison}(d) illustrates the average training time per epoch, where PD-Loss Random T1.0 remains efficient at 16.75 seconds, comparable to Circle, Proxy NCA, and Proxy Anchor losses, and significantly faster than MS loss, which requires 21.93 seconds per epoch.
\begin{figure}[!ht]
    \centering
        \begin{tabular}{@{}c@{}c@{}}
            \includegraphics[width=3.2in,height=2.1in]{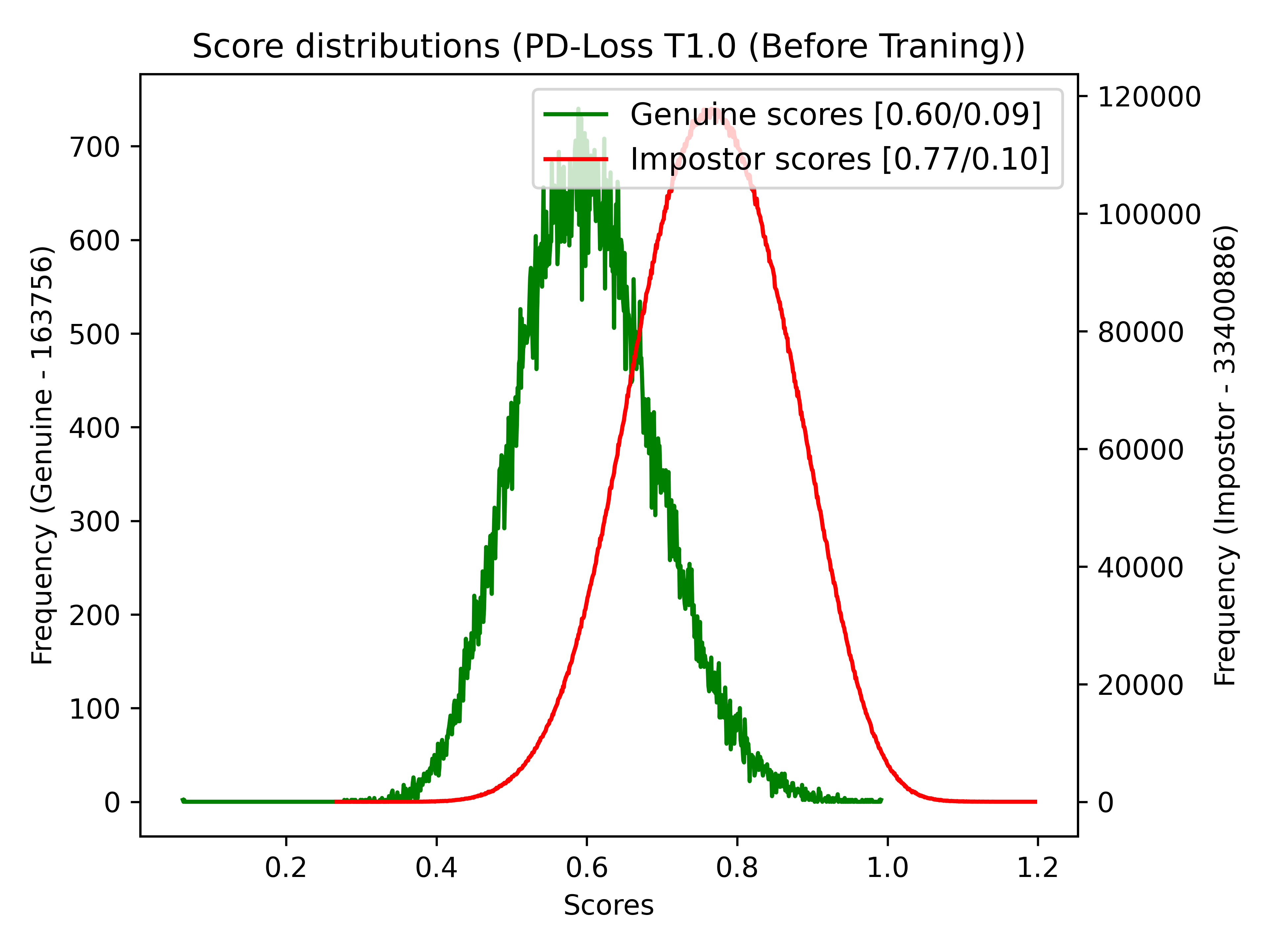}
            & \includegraphics[width=3.2in,height=2.1in]{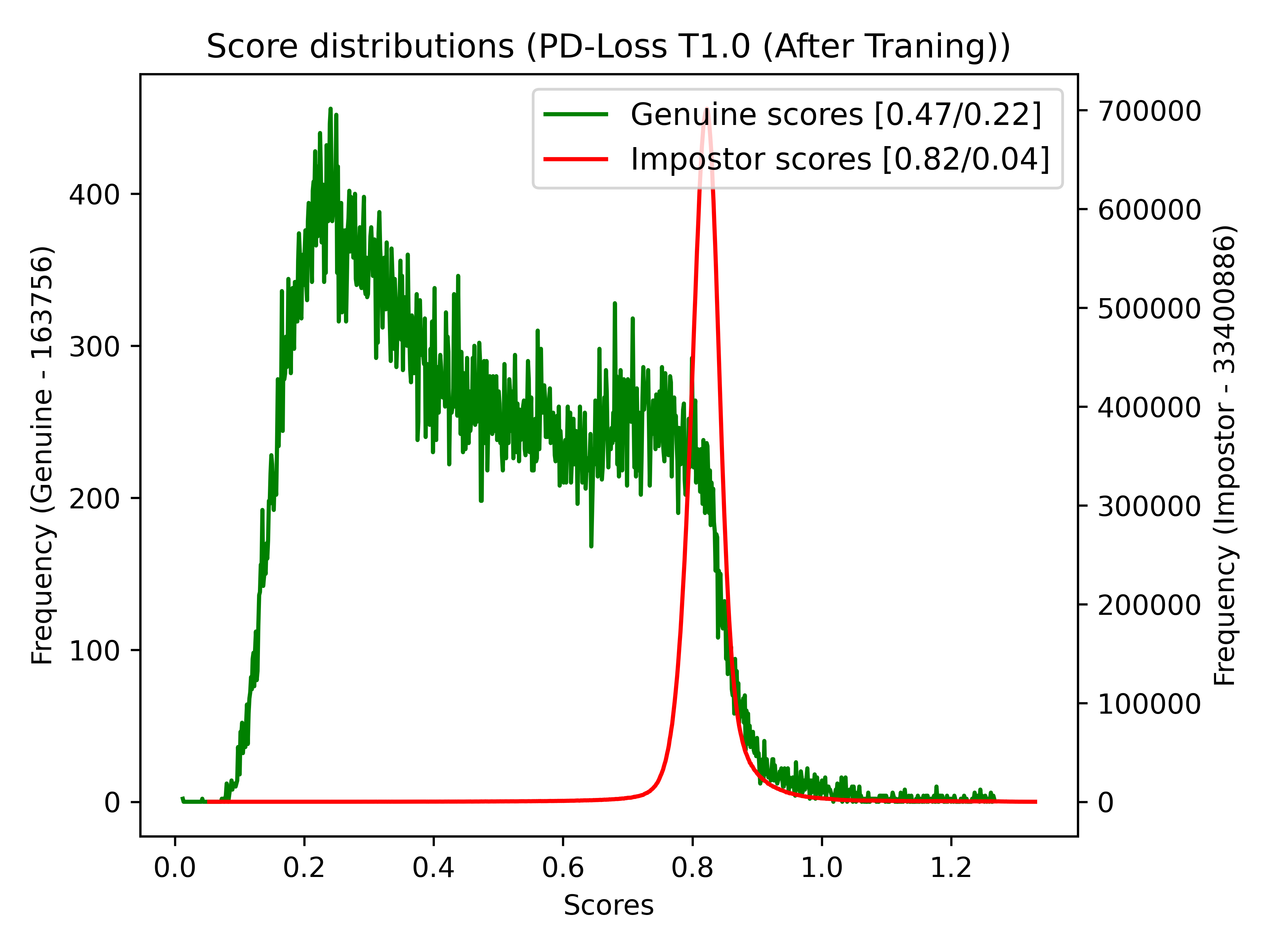}  \\
            (a) & (b)  
        \end{tabular}
    \caption{Distributions of genuine and impostor distances on the CUB-200 test set, before (a) and after (b) training with PD-Loss ($\tau$=1.0), EMB 512, and BS 32.}
    \label{fig:analysis-v2}
\end{figure}

Figures~\ref{fig:analysis-v2}(a) and \ref{fig:analysis-v2}(b) illustrate the genuine and impostor distance distributions for the CUB-200 test set embeddings, comparing the model before and after training with PD-Loss (Random Init $\tau=1.0$). Before training (Figure~\ref{fig:analysis-v2}(a)), the genuine (green) and impostor (red) distributions exhibit significant overlap, with the genuine mean (0.60) close to the impostor mean (0.77), resulting in a low $d'$ (0.91). After training (Figure~\ref{fig:analysis-v2}(b)), the distributions are markedly separated. The genuine distribution shifts towards lower distances (mean 0.47) and becomes more compact (smaller standard deviation, std 0.22), while the impostor distribution shifts towards higher distances (mean 0.82, std 0.04). This clear separation is reflected in a substantially increased $d'$ value (2.19). This visual evidence strongly supports the effectiveness of the PD-Loss objective in optimizing the Decidability Index to achieve distinct statistical separation between genuine and impostor pairs in the embedding space.

\section{Conclusion}
\label{sec:conclusion}

In this work, we introduced Proxy-Decidability Loss (PD-Loss), a novel deep metric learning objective. Unlike traditional methods that rely on direct sample-to-sample or sample-to-proxy comparisons, PD-Loss optimizes the global statistical separability (Decidability Index, $d'$) of genuine and impostor distributions. It achieves this efficiently by using learnable proxies to estimate these distributions within mini-batches, thereby inheriting scalability benefits without requiring direct data-to-data interactions for loss calculation. Enhanced by a logarithmic formulation and temperature scaling, PD-Loss demonstrates competitive or state-of-the-art performance on CUB-200, LFW, and CARS196 datasets. Our findings highlight the benefits of random proxy initialization and high temperature.

\section*{Acknowledgment}
This work was supported by the Conselho Nacional de Desenvolvimento Científico e Tecnológico (CNPq, grants 308400/2022-4, 307151/2022-0), Coordenação de Aperfeiçoamento de Pessoal de Nível Superior (CAPES - grant 001), Fundação de Amparo à Pesquisa do Estado de Minas Gerais (FAPEMIG, grant APQ-01647-22). We also thank the Universidade Federal de Ouro Preto (UFOP) for their support.





\bibliographystyle{main}  
\bibliography{main}

\begin{thebibliography}{10}
\expandafter\ifx\csname url\endcsname\relax
  \def\url#1{\texttt{#1}}\fi
\expandafter\ifx\csname urlprefix\endcsname\relax\def\urlprefix{URL }\fi
\expandafter\ifx\csname href\endcsname\relax
  \def\href#1#2{#2} \def\path#1{#1}\fi

\bibitem{gordo2016deep}
A.~Gordo, J.~Almaz{\'a}n, J.~Revaud, D.~Larlus, Deep image retrieval: Learning global representations for image search, in: 14th European Conference on Computer Vision (ECCV 2016), 2016, pp. 241--257.

\bibitem{schroff_facenet_15}
F.~Schroff, D.~Kalenichenko, J.~Philbin, Facenet: A unified embedding for face recognition and clustering, in: Proceedings of the IEEE conference on computer vision and pattern recognition, 2015, pp. 815--823.

\bibitem{hermans2017defense}
A.~Hermans, L.~Beyer, B.~Leibe, In defense of the triplet loss for person re-identification, arXiv preprint arXiv:1703.07737.

\bibitem{snell2017prototypical}
J.~Snell, K.~Swersky, R.~Zemel, Prototypical networks for few-shot learning, Advances in Neural Information Processing Systems 30.

\bibitem{hadsell2006dimensionality}
R.~Hadsell, S.~Chopra, Y.~LeCun, Dimensionality reduction by learning an invariant mapping, in: IEEE Conference on Computer Vision and Pattern Recognition (CVPR'06), Vol.~2, IEEE, 2006, pp. 1735--1742.

\bibitem{sohn2016improved}
K.~Sohn, Improved deep metric learning with multi-class n-pair loss objective, Advances in Neural Information Processing Systems 29.

\bibitem{wang_msloss_19}
X.~Wang, X.~Han, W.~Huang, D.~Dong, M.~R. Scott, Multi-similarity loss with general pair weighting for deep metric learning, in: Proceedings of the IEEE/CVF conference on computer vision and pattern recognition, 2019, pp. 5022--5030.

\bibitem{sun_circleloss_20}
Y.~Sun, C.~Cheng, Y.~Zhang, C.~Zhang, L.~Zheng, Z.~Wang, Y.~Wei, Circle loss: A unified perspective of pair similarity optimization, in: Proceedings of the IEEE/CVF conference on computer vision and pattern recognition, 2020, pp. 6398--6407.

\bibitem{wu2017sampling}
C.-Y. Wu, R.~Manmatha, A.~J. Smola, P.~Krahenbuhl, Sampling matters in deep embedding learning, in: Proceedings of the IEEE international conference on computer vision, 2017, pp. 2840--2848.

\bibitem{movshovitz2017no}
Y.~Movshovitz-Attias, A.~Toshev, T.~K. Leung, S.~Ioffe, S.~Singh, No fuss distance metric learning using proxies, in: Proceedings of the IEEE international conference on computer vision, 2017, pp. 360--368.

\bibitem{kim2020proxy}
S.~Kim, D.~Kim, M.~Cho, S.~Kwak, Proxy anchor loss for deep metric learning, in: Proceedings of the IEEE/CVF conference on computer vision and pattern recognition, 2020, pp. 3238--3247.

\bibitem{SilvaIJCNN22}
P.~H. Silva, G.~Moreira, V.~Freitas, R.~Silva, D.~Menotti, E.~Luz, A decidability-based loss function, in: 2022 International Joint Conference on Neural Networks (IJCNN), IEEE, 2022, pp. 1--8.

\bibitem{daugman2000biometric}
J.~Daugman, Biometric decision landscapes, Tech. rep., University of Cambridge, Computer Laboratory (2000).

\bibitem{wah2011caltech}
C.~Wah, S.~Branson, P.~Welinder, P.~Perona, S.~Belongie, The caltech-ucsd birds-200-2011 dataset (2011).

\bibitem{krause20133d}
J.~Krause, M.~Stark, J.~Deng, L.~Fei-Fei, 3d object representations for fine-grained categorization, in: Proceedings of the IEEE international conference on computer vision workshops, 2013, pp. 554--561.

\bibitem{LFWTechUpdate}
G.~B. Huang, M.~Mattar, T.~Berg, E.~Learned-Miller, Labeled faces in the wild: A database forstudying face recognition in unconstrained environments, in: Workshop on faces in'Real-Life'Images: detection, alignment, and recognition, 2008.

\bibitem{loshchilov2017decoupled}
I.~Loshchilov, F.~Hutter, Decoupled weight decay regularization, arXiv preprint arXiv:1711.05101.

\bibitem{paszke2019pytorch}
A.~Paszke, Pytorch: An imperative style, high-performance deep learning library, arXiv preprint arXiv:1912.01703.

\bibitem{he2016deep}
K.~He, X.~Zhang, S.~Ren, J.~Sun, Deep residual learning for image recognition, in: Proceedings of the IEEE conference on computer vision and pattern recognition, 2016, pp. 770--778.

\bibitem{musgrave2020pytorch}
K.~Musgrave, S.~Belongie, S.-N. Lim, Pytorch metric learning, arXiv preprint arXiv:2008.09164.

\end{thebibliography}

\end{document}